\documentclass{article}

\usepackage[final]{IIBPSL_2020}
\usepackage[utf8]{inputenc}
\usepackage{graphicx}
\usepackage{float}
\usepackage[ruled, vlined]{algorithm2e}
\usepackage{subfig}
\usepackage{color}
\usepackage{fullpage}
\usepackage{mkolar_definitions}
\usepackage{commath}
\usepackage{hyperref}
\DeclareMathOperator{\ce}{CE}

\renewcommand{\paragraph}[1]{\vspace{1mm} \noindent{\bf #1}}

\title{Improving the trustworthiness of image classification models by utilizing bounding-box annotations}

\author{Dharma KC\\
University of Arizona\\
\texttt{kcdharma@email.arizona.edu}
\And 
Chicheng Zhang \\
University of Arizona\\
\texttt{chichengz@cs.arizona.edu}
}
\newcommand{\lambdavary}{\ensuremath{\lambda\textsc{-vary}}\xspace}
\newcommand{\lambdaequal}{\ensuremath{\lambda\textsc{-equal}}\xspace}

\newcommand{\blackout}{\ensuremath{\textsc{Blackout}}\xspace}
\newcommand{\standard}{\ensuremath{\textsc{Standard}}\xspace}

\def\shownotes{1}  \ifnum\shownotes=1
\newcommand{\authnote}[2]{$\ll$\textsf{\footnotesize #1 notes: #2}$\gg$}
\else
 \newcommand{\authnote}[2]{}
\fi

\begin{document}

\maketitle

\begin{abstract}
We study utilizing auxiliary information in training data to improve the trustworthiness of machine learning models. Specifically, in the context of image classification, we propose to optimize a training objective that incorporates  bounding box information, which is available in many image classification datasets. Preliminary experimental results show that the proposed algorithm achieves better performance in accuracy, robustness, and interpretability compared with baselines.
\end{abstract}

\section{Introduction}
Building reliable and trustworthy prediction models has long been a central topic of machine learning research. The reliability and trustworthiness of machine learning models can be characterized from many aspects, including test accuracy, robustness against adversarial attacks~\cite{goodfellow2014explaining}, interpretability~\cite{molnar2019}, etc. To earn trust and adoption from human users, it is important to develop machine learning models that have good performance on all these aspects.

However, many works show that there might be fundamental tradeoffs between these performance metrics. For example, it is shown by~\cite{tsipras2018robustness} that adversarial robustness may be at odds with accuracy.  
As another example, decision trees or sparse linear models enjoy global interpretability, however their expressivity may be limited~\cite{craven1996extracting,shalev2010trading}.

On the other hand, in many real-world supervised machine learning applications, rich auxiliary information beyond (feature, label) pairs is available at training time. For example, many object detection benchmark datasets provide bounding-box annotations for  images~\cite{su2012crowdsourcing,lin2014microsoft}. 
As another example, in ECG-based heart disease prediction, doctors can highlight parts of signals most indicative of her diagnoses.
This motivates the question: can learning algorithms benefit from these auxiliary information to train models with improved test accuracy, adversarial robustness and interpretability?

There have been considerable efforts addressing the above question, with most progress focusing on improving test accuracy (e.g. \cite{vapnik2009new,lambert2018deep,lopez2015unifying}). 
Although there are some recent efforts aiming to build models with improved accuracy and interpretability (e.g. \cite{ross2017right, mitsuhara2019embedding, rieger2019interpretations}),
a comprehensive understanding of when and how learning from auxiliary information help improve other  aspects of trustworthiness of machine learning models (e.g. adversarial robustness) is still missing.

In this paper, we study learning from auxiliary information with the goal of simultaneously improving accuracy, adversarial robustness and interpretability. Specifically, we focus on image classification, and consider auxiliary information in the form of object bounding boxes~\cite{su2012crowdsourcing,lin2014microsoft}. Inspired by works on gradient-based regularization~\cite{drucker1992improving,ross2017right,ross2018improving}, we propose a training objective that has different degrees of regularization on different parts of input data, taking advantage of bounding box auxiliary information. Experimentally, we demonstrate on the Caltech-UCSD Birds dataset~\cite{wah2011caltech} that our proposed algorithm outputs image classification models with improved accuracy, robustness and interpretability, both quantitatively and qualitatively. Our open source code is available at: \href{https://github.com/ck-amrahd/birds}{https://github.com/ck-amrahd/birds}.

\section{Related work}

Advanced model training algorithms aiming at improving adversarial robustness have been proposed in the literature~\cite{szegedy2013intriguing,drucker1992improving,ross2018improving}. However, improvements on robust accuracy often comes at the price of lower standard accuracy~\cite{tsipras2018robustness}.

Learning from auxiliary information beyond labels has been studied in various contexts in the literature, for example in text \cite{zaidan2008machine, sharma2018learning, zhang2016rationale}
and image \cite{donahue2011annotator} domains. 
It has also received more theoretical treatments in the works of \cite{vapnik2009new, poulis2017learning, dasgupta2018learning}. 
In the context of image classification, several works aim at improving model accuracy and interpretability using bounding box-based auxiliary information. 
\cite{mitsuhara2019embedding} penalizes the mismatch between the model-generated attention masks and bounding boxes to improve the accuracy and interpratability of convolutional neural networks (CNNs). \cite{ross2017right} proposes a regularization term in the training objective that penalizes the gradients of cross entropy losses with respect to input features outside bounding boxes; our work can be seen as a generalization of that work, in that we additionally incorporate gradient-based penalty with respect to the features inside bounding boxes.
\cite{zhang2014part} utilizes more refined part localization bounding box information to train CNNs and improves model accuracy in fine grained classification tasks. Recently, \cite{rieger2019interpretations} utilizes the attribution algorithm of \cite{singh2018hierarchical} to propose a new regularizer that can flexibly encourage or penalize different parts of input features.

Our work is also closely related to attribution map or saliency map generation for images~\cite{simonyan2013deep,zhou2016learning,selvaraju2017grad,dabkowski2017real}, in that one can propose training objectives that promote ``alignment'' between such attribution maps and bounding boxes.
Although our work only focuses on regularization based on gradient-based explanations,
we believe that regularizing based on these more sophisticated attribution maps are interesting avenues towards improving the trustworthiness of image classification models.

Finally, recent works empirically demonstrate that adversarial robustness and interpretability, two important performance metrics considered in this paper, are tightly connected. 
On one hand, adversarially robust models generate more interpretable explanations than non-robust ones \cite{tsipras2018robustness, etmann2019connection, kim2019bridging}; on the other hand, models trained to mimic gradient-based explanations of adversarially robust models exhibit robustness~\cite{noack2019does}, hinting at the possibility that robustness is a side-benefit of interpretability.

\section{Algorithm}

\begin{minipage}{0.65\textwidth}
\paragraph{Definitions and settings.} We study image classification with bounding box annotations being part of training data.
We are given a set of $m$ training examples $\cbr{(x_i, y_i, M_i)}_{i=1}^m$, where for example $i$, $x_i \in \RR^d$ is its feature part (image $i$'s pixel representation), $y_i \in [K]$ is its label part (the class of the object in the image), $M_i \subseteq [d]$ is the image's associated bounding box. An example of an image with bounding box information is given in Figure~\ref{fig:ground_truth}. 
Our goal is to train a neural network-based classification model such that, when predicting on test examples, it has high accuracy, robustness and good interpretability. 

Formally, given an example $x$, our network outputs a prediction $f(x; \theta)$ that is a probability vector in $\Delta^{K-1}$, the $K$-dimensional probability simplex. 
Define the cross entropy loss of model $f(\cdot; \theta)$ on example $(x,y)$ as 
$\ell_{\ce}(\theta, (x,y)) \triangleq \ln\frac{1}{f^y(x; \theta)}$, where we use the notation $z^j$ to denote the $j$-th coordinate of vector $z$.  
\end{minipage}
\hspace{0.2cm}
\begin{minipage}{0.32\textwidth}
\begin{figure}[H]
    \centering
    \includegraphics[height=4cm]{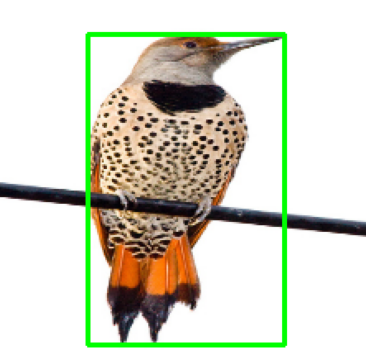}
    \caption{A Northern flicker image with bounding box shown in green, taken from~\cite{wah2011caltech}.}
    \label{fig:ground_truth}
\end{figure}
\end{minipage}

\paragraph{Training objective.} For model training, we propose to optimize the following objective function: $\min_\theta \sum_{i=1}^m \ell(\theta, (x_i, y_i, M_i))$, where 
\begin{equation} 
\ell(\theta, (x, y, M) ) \triangleq \ell_{\ce}(\theta, (x,y))
+ 
\lambda_1 \sum_{j \in M} \rbr{ \pd{\ell_{\ce}(\theta, (x,y))}{{x^j}} }^2
+ 
\lambda_2 \sum_{j \in [d] \setminus M} \rbr{ \pd{\ell_{\ce}(\theta, (x,y))}{{x^j}} }^2
\label{eqn:objective}
\end{equation}
for some $\lambda_1, \lambda_2 > 0$.
The intuition behind this training objective is that, 
in addition to minimizing the usual cross entropy loss, 
we would like to ensure that the model's predictions have different degrees of sensitivity to different parts of the training images. Specifically, the magnitude of $\pd{\ell_{\ce}(\theta, (x,y))}{{x^j}}$ characterizes the sensitivity of the cross entropy loss with respect to the $j$-th pixel. We would like to train a model whose sensitivity to input aligns with object bounding boxes as much as possible; formally, $\pd{\ell_{\ce}(\theta, (x,y))}{{x^j}}$ should be large for $j$ in $M$ and should be small otherwise.

\paragraph{Comparison to prior works.} Our training objective generalizes those of~\cite{drucker1992improving, ross2017right, ross2018improving}. Specifically, when $\lambda_1 = 0$, the objective function becomes that of~\cite{ross2017right}, where only the sensitivity of the loss with respect to the irrelevant parts of the input (coordinates in $[d] \setminus M$) are penalized; \cite{ross2017right} shows that this formulation promotes model interpretability. On the other hand, when $\lambda_1 = \lambda_2$, the objective function becomes the double backpropagation objective~\cite{drucker1992improving, ross2018improving}, which is known to improve the generalization accuracy and adversarial robustness of models.  

\section{Experiments}
We use the Caltech-UCSD Birds (abbrev. CUB) dataset \cite{wah2011caltech} for experimental evaluation, which has 11,788 examples. 
We take the union of the training and test sets provided by the CUB dataset, permute the set, and perform a four-way split. 
The first split consists of $1/2$ of the data, which is used for training by optimizing our objective~\eqref{eqn:objective}. The remaining data is divided into three sets of equal sizes: the first set is used to select the best model during training, the second set is for $\lambda_1$ and $\lambda_2$ hyperparameter selection, and the third set is used for testing.
We choose ResNet50 \cite{he2016deep} as our model architecture, and train with mini-batch stochastic gradient descent with a learning rate of 0.001.
We consider training with the choices of $\lambda_1$ and $\lambda_2$ in a grid $\Lambda^2$, where $\Lambda = \cbr{0} \cup \cbr{(\sqrt[3]{10})^i: i \in \cbr{-3, -2, \ldots, 9}}$. 
All experiments are repeated three times. 
We evaluate the following set of algorithms:
\begin{enumerate}
    \item \lambdavary (our proposed approach): train a model for each $(\lambda_1, \lambda_2)$ in $\Lambda^2$, and use the validation set to select the best performing model. %
    \item \lambdaequal: train a model for each $(\lambda_1, \lambda_2)$ in $\cbr{(\lambda_1, \lambda_2) \in \Lambda^2: \lambda_1 = \lambda_2}$, and use the validation set to select the best performing model.
    \item \blackout: train a model
    that minimizes the cross entropy loss over modified training examples $(\tilde{x}_i, y_i)$'s; here $\tilde{x}_i$ is defined as $x_i$ with coordinates outside $M_i$ set to zero. 
    \item \standard: standard training that minimizes the cross entropy loss over $(x_i, y_i)$'s; this is also equivalent to setting $\lambda_1 = \lambda_2 = 0$.
\end{enumerate}

\subsection{Standard and robust accuracy comparison}

\begin{minipage}{0.6\textwidth}
The adversarial robustness of our trained models are tested for 10 values of adversarial perturbation radii $\epsilon$'s in $\cbr{\frac{0.2 i}{9}: i \in \cbr{0,\ldots,9}}$ using the Fast Gradient Sign Method \cite{goodfellow2014explaining}. We choose max value of $\epsilon$ to be $0.2$ because beyond that perturbation, the images become unrecognizable by humans. Our adversarial examples were generated using the Foolbox library \cite{rauber2017foolbox}.
Recall that for \lambdavary and \lambdaequal, for each value of $\epsilon$, we choose separate values of $(\lambda_1, \lambda_2)$ pairs using the validation set. 

Our results are shown in Figure~\ref{fig:robustness}. It can be seen that \lambdavary trains models that have higher standard  accuracy and also robust to adversarial attacks; the performance of the learned models beat those of \lambdaequal (especially when $\epsilon$ is large), showing the utility of incorporating bounding box information in the training objective. 
\end{minipage}
\hspace{0.2cm}
\begin{minipage}{0.4\textwidth}
\vspace{-1cm}
\begin{figure}[H]
    \centering
    \includegraphics[width=2.4in]{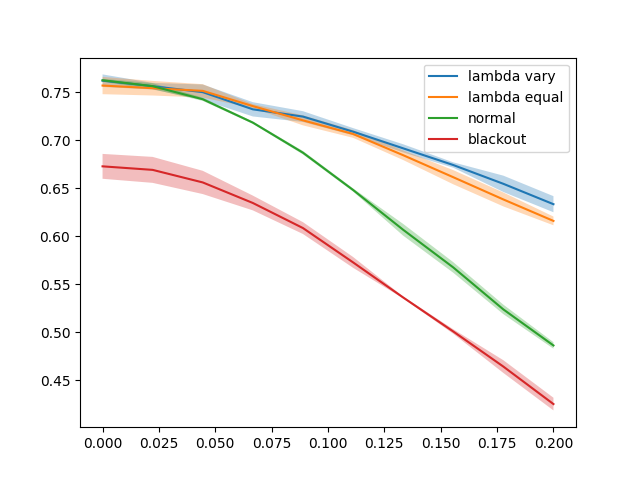}
    \caption{Test robust accuracy for different values of $\epsilon$'s, for the CUB dataset; the error bands here represent standard deviation.}
    \label{fig:robustness}
\end{figure}
\end{minipage}

\subsection{Interpretability comparison}
We compare the interpretability of the trained models qualitatively and quantitatively. 

\begin{figure}
    \subfloat[original]{\includegraphics[width=1in, height=1in]{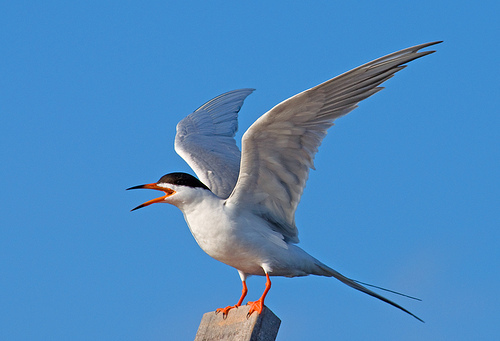}}
    \hspace{0.1cm}
    \subfloat[]{\includegraphics[width=1in, height=1in]{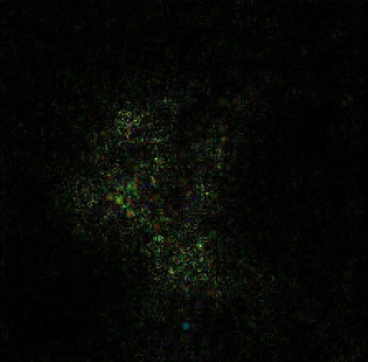}}
    \hspace{0.1cm}
    \subfloat[]{\includegraphics[width=1in, height=1in]{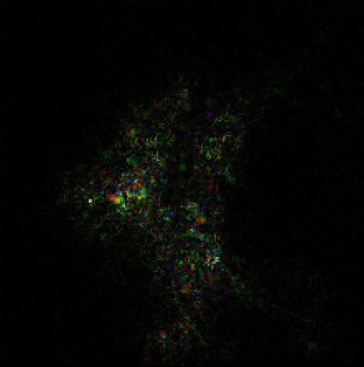}}
    \hspace{0.1cm}
    \subfloat[]{\includegraphics[width=1in, height=1in]{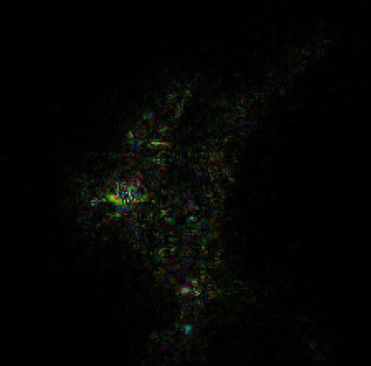}}
    \hspace{0.1cm}
    \subfloat[]{\includegraphics[width=1in, height=1in]{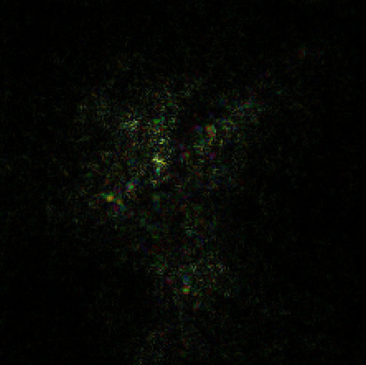}}
    \caption{Gradient-based saliency maps of a bird image (a) generated by models trained using different objectives: (b) \standard; (c) \lambdaequal; (d) \lambdavary; (e) \blackout. %
    }
    \label{fig:bird-saliency-1}
    \vspace{-0.5cm}
\end{figure}

\paragraph{Qualitative results.} 
We plot the gradient-based saliency map ~\cite{simonyan2013deep} 
generated by the model trained by each algorithm on a few bird images in the CUB dataset.  We can see from Figure~\ref{fig:bird-saliency-1} that the saliency map of the \standard and \blackout is dispersed and clearly not focusing on the bird body. The saliency map of \lambdaequal is doing better than \standard and finally the model trained by \lambdavary even highlights subtle parts such as beaks and legs with the complete shape of bird. 

\begin{figure}
\subfloat[]{\includegraphics[width=1.3in]{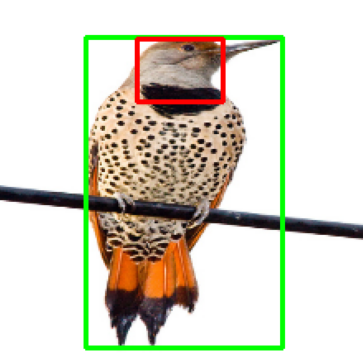}}
\subfloat[]{\includegraphics[width=1.3in]{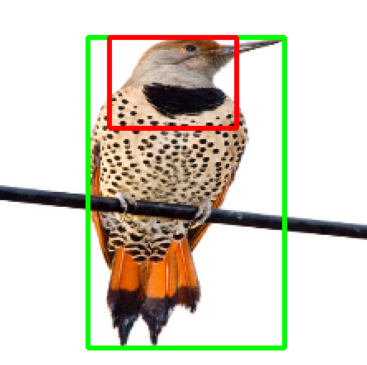}}
\subfloat[]{\includegraphics[width=1.3in]{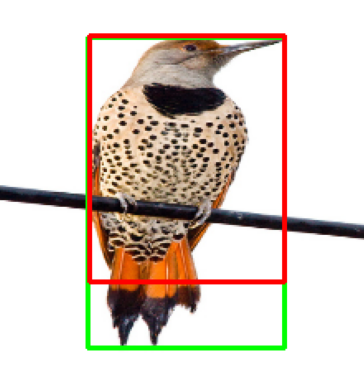}}
\subfloat[]{\includegraphics[width=1.3in]{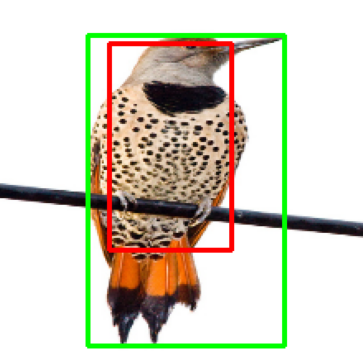}}
\caption{Localization results on the bird image in Figure~\ref{fig:ground_truth}, generated by models trained using different objectives: (a) \standard; (b) \lambdaequal; (c) \lambdavary; (d) \blackout. Here for each model, we extract a bounding box (shown in red) from its gradient-based saliency map.}
\label{fig:bounding-box}
\vspace{-0.5cm}
\end{figure}

\paragraph{Quantitative results.} To quantitatively measure the interpretability of the gradient-based saliency maps output by different models, we extract bounding boxes from them and evaluate them in two ways: first, we use the saliency metric proposed in~\cite{dabkowski2017real}; second, we compare the extracted bounding boxes with the ground truth bounding boxes using localization accuracy \cite{Everingham10}.

To generate a bounding box from a saliency map, we binarize the image by thresholding, and output the tightest rectangular box that contains the pixels whose grayscale is above the threshold.

\paragraph{Saliency metric:}
We follow~\cite{dabkowski2017real} to perform the following calculation: after generating a bounding box, crop the corresponding region from the original image and pass it into the network to make prediction. The saliency metric of~\cite{dabkowski2017real} is defined as: $s(a, p) = \log(a) - \log(p)$, where $a = \max(0.05, \hat{a})$, and $\hat{a}$ is the area fraction of the bounding box, and $p$ is the model's predictive probability for the correct label. The lower value the saliency metric the better. Table~\ref{tab:saliency-metric} shows the test saliency metric of models trained by all methods, where we can see that \lambdavary outperforms the baselines.

\begin{table}[H]
    \centering
    \begin{tabular}{|c|c|c|c|}
    \hline
    \standard & \blackout & \lambdaequal & \lambdavary \\
    \hline
     0.466 $\pm$ 0.047 & 0.396 $\pm$ 0.033 & 0.343 $\pm$ 0.02 & \textbf{0.283 $\pm$ 0.03} \\
    \hline
    \end{tabular}
    \vspace{0.1cm}
    \caption{Saliency metric comparison among the evaluated methods on the CUB dataset.}
    \label{tab:saliency-metric}
\end{table}

\paragraph{Localization accuracy:}
The localization accuracy is defined as the fraction of examples where the model prediction is correct and the generated bounding box has intersection over union (IOU) value of $\geq 0.5$ with the ground truth bounding box. Table~\ref{tab:saliency-metric} shows the test localization accuracy of models trained by all methods, where we can see that \lambdavary outperforms the baselines.

\begin{table}[H]
    \centering
    \begin{tabular}{|c|c|c|c|}
    \hline
    \standard & \blackout & \lambdaequal & \lambdavary \\
    \hline
    0.236 $\pm$ 0.02 & 0.30 $\pm$ 0.021 & 0.30 $\pm$ 0.169 & \textbf{0.343 $\pm$ 0.012} \\
    \hline
    \end{tabular}
    \vspace{0.1cm}
    \caption{Localization accuracy comparison among the evaluated methods on the CUB dataset.}
    \label{tab:localization}
\end{table}

\bibliographystyle{plain}
\bibliography{refs}

\begin{thebibliography}{10}

\bibitem{craven1996extracting}
Mark Craven and Jude~W Shavlik.
\newblock Extracting tree-structured representations of trained networks.
\newblock In {\em Advances in neural information processing systems}, pages
  24--30, 1996.

\bibitem{dabkowski2017real}
Piotr Dabkowski and Yarin Gal.
\newblock Real time image saliency for black box classifiers.
\newblock In {\em Advances in Neural Information Processing Systems}, pages
  6967--6976, 2017.

\bibitem{dasgupta2018learning}
Sanjoy Dasgupta, Akansha Dey, Nicholas Roberts, and Sivan Sabato.
\newblock Learning from discriminative feature feedback.
\newblock In {\em Advances in Neural Information Processing Systems}, pages
  3955--3963, 2018.

\bibitem{donahue2011annotator}
Jeff Donahue and Kristen Grauman.
\newblock Annotator rationales for visual recognition.
\newblock In {\em 2011 International Conference on Computer Vision}, pages
  1395--1402. IEEE, 2011.

\bibitem{drucker1992improving}
Harris Drucker and Yann Le~Cun.
\newblock Improving generalization performance using double backpropagation.
\newblock {\em IEEE Transactions on Neural Networks}, 3(6):991--997, 1992.

\bibitem{etmann2019connection}
Christian Etmann, Sebastian Lunz, Peter Maass, and Carola-Bibiane
  Sch{\"o}nlieb.
\newblock On the connection between adversarial robustness and saliency map
  interpretability.
\newblock {\em arXiv preprint arXiv:1905.04172}, 2019.

\bibitem{Everingham10}
M.~Everingham, L.~Van~Gool, C.~K.~I. Williams, J.~Winn, and A.~Zisserman.
\newblock The pascal visual object classes (voc) challenge.
\newblock {\em International Journal of Computer Vision}, 88(2):303--338, June
  2010.

\bibitem{goodfellow2014explaining}
Ian~J Goodfellow, Jonathon Shlens, and Christian Szegedy.
\newblock Explaining and harnessing adversarial examples.
\newblock {\em arXiv preprint arXiv:1412.6572}, 2014.

\bibitem{he2016deep}
Kaiming He, Xiangyu Zhang, Shaoqing Ren, and Jian Sun.
\newblock Deep residual learning for image recognition.
\newblock In {\em Proceedings of the IEEE conference on computer vision and
  pattern recognition}, pages 770--778, 2016.

\bibitem{kim2019bridging}
Beomsu Kim, Junghoon Seo, and Taegyun Jeon.
\newblock Bridging adversarial robustness and gradient interpretability.
\newblock {\em arXiv preprint arXiv:1903.11626}, 2019.

\bibitem{lambert2018deep}
John Lambert, Ozan Sener, and Silvio Savarese.
\newblock Deep learning under privileged information using heteroscedastic
  dropout.
\newblock In {\em Proceedings of the IEEE Conference on Computer Vision and
  Pattern Recognition}, pages 8886--8895, 2018.

\bibitem{lin2014microsoft}
Tsung-Yi Lin, Michael Maire, Serge Belongie, James Hays, Pietro Perona, Deva
  Ramanan, Piotr Doll{\'a}r, and C~Lawrence Zitnick.
\newblock Microsoft coco: Common objects in context.
\newblock In {\em European conference on computer vision}, pages 740--755.
  Springer, 2014.

\bibitem{lopez2015unifying}
David Lopez-Paz, L{\'e}on Bottou, Bernhard Sch{\"o}lkopf, and Vladimir Vapnik.
\newblock Unifying distillation and privileged information.
\newblock {\em arXiv preprint arXiv:1511.03643}, 2015.

\bibitem{mitsuhara2019embedding}
Masahiro Mitsuhara, Hiroshi Fukui, Yusuke Sakashita, Takanori Ogata, Tsubasa
  Hirakawa, Takayoshi Yamashita, and Hironobu Fujiyoshi.
\newblock Embedding human knowledge in deep neural network via attention map.
\newblock {\em arXiv preprint arXiv:1905.03540}, 5, 2019.

\bibitem{molnar2019}
Christoph Molnar.
\newblock {\em Interpretable Machine Learning}.
\newblock 2019.
\newblock \url{https://christophm.github.io/interpretable-ml-book/}.

\bibitem{noack2019does}
Adam Noack, Isaac Ahern, Dejing Dou, and Boyang Li.
\newblock Does interpretability of neural networks imply adversarial
  robustness?
\newblock {\em arXiv preprint arXiv:1912.03430}, 2019.

\bibitem{poulis2017learning}
Stefanos Poulis and Sanjoy Dasgupta.
\newblock Learning with feature feedback: from theory to practice.
\newblock In {\em Artificial Intelligence and Statistics}, pages 1104--1113,
  2017.

\bibitem{rauber2017foolbox}
Jonas Rauber, Wieland Brendel, and Matthias Bethge.
\newblock Foolbox: A python toolbox to benchmark the robustness of machine
  learning models.
\newblock In {\em Reliable Machine Learning in the Wild Workshop, 34th
  International Conference on Machine Learning}, 2017.

\bibitem{rieger2019interpretations}
Laura Rieger, Chandan Singh, W~James Murdoch, and Bin Yu.
\newblock Interpretations are useful: penalizing explanations to align neural
  networks with prior knowledge.
\newblock {\em arXiv preprint arXiv:1909.13584}, 2019.

\bibitem{ross2018improving}
Andrew~Slavin Ross and Finale Doshi-Velez.
\newblock Improving the adversarial robustness and interpretability of deep
  neural networks by regularizing their input gradients.
\newblock In {\em Thirty-second AAAI conference on artificial intelligence},
  2018.

\bibitem{ross2017right}
Andrew~Slavin Ross, Michael~C Hughes, and Finale Doshi-Velez.
\newblock Right for the right reasons: Training differentiable models by
  constraining their explanations.
\newblock {\em arXiv preprint arXiv:1703.03717}, 2017.

\bibitem{selvaraju2017grad}
Ramprasaath~R Selvaraju, Michael Cogswell, Abhishek Das, Ramakrishna Vedantam,
  Devi Parikh, and Dhruv Batra.
\newblock Grad-cam: Visual explanations from deep networks via gradient-based
  localization.
\newblock In {\em Proceedings of the IEEE international conference on computer
  vision}, pages 618--626, 2017.

\bibitem{shalev2010trading}
Shai Shalev-Shwartz, Nathan Srebro, and Tong Zhang.
\newblock Trading accuracy for sparsity in optimization problems with sparsity
  constraints.
\newblock {\em SIAM Journal on Optimization}, 20(6):2807--2832, 2010.

\bibitem{sharma2018learning}
Manali Sharma and Mustafa Bilgic.
\newblock Learning with rationales for document classification.
\newblock {\em Machine Learning}, 107(5):797--824, 2018.

\bibitem{simonyan2013deep}
Karen Simonyan, Andrea Vedaldi, and Andrew Zisserman.
\newblock Deep inside convolutional networks: Visualising image classification
  models and saliency maps.
\newblock {\em arXiv preprint arXiv:1312.6034}, 2013.

\bibitem{singh2018hierarchical}
Chandan Singh, W~James Murdoch, and Bin Yu.
\newblock Hierarchical interpretations for neural network predictions.
\newblock In {\em International Conference on Learning Representations}, 2018.

\bibitem{su2012crowdsourcing}
Hao Su, Jia Deng, and Li~Fei-Fei.
\newblock Crowdsourcing annotations for visual object detection.
\newblock In {\em Workshops at the Twenty-Sixth AAAI Conference on Artificial
  Intelligence}, 2012.

\bibitem{szegedy2013intriguing}
Christian Szegedy, Wojciech Zaremba, Ilya Sutskever, Joan Bruna, Dumitru Erhan,
  Ian Goodfellow, and Rob Fergus.
\newblock Intriguing properties of neural networks.
\newblock {\em arXiv preprint arXiv:1312.6199}, 2013.

\bibitem{tsipras2018robustness}
Dimitris Tsipras, Shibani Santurkar, Logan Engstrom, Alexander Turner, and
  Aleksander Madry.
\newblock Robustness may be at odds with accuracy.
\newblock {\em arXiv preprint arXiv:1805.12152}, 2018.

\bibitem{vapnik2009new}
Vladimir Vapnik and Akshay Vashist.
\newblock A new learning paradigm: Learning using privileged information.
\newblock {\em Neural networks}, 22(5-6):544--557, 2009.

\bibitem{wah2011caltech}
Catherine Wah, Steve Branson, Peter Welinder, Pietro Perona, and Serge
  Belongie.
\newblock The caltech-ucsd birds-200-2011 dataset.
\newblock 2011.

\bibitem{zaidan2008machine}
Omar~F Zaidan, Jason Eisner, and Christine Piatko.
\newblock Machine learning with annotator rationales to reduce annotation cost.
\newblock In {\em Proceedings of the NIPS* 2008 workshop on cost sensitive
  learning}, pages 260--267, 2008.

\bibitem{zhang2014part}
Ning Zhang, Jeff Donahue, Ross Girshick, and Trevor Darrell.
\newblock Part-based r-cnns for fine-grained category detection.
\newblock In {\em European conference on computer vision}, pages 834--849.
  Springer, 2014.

\bibitem{zhang2016rationale}
Ye~Zhang, Iain Marshall, and Byron~C Wallace.
\newblock Rationale-augmented convolutional neural networks for text
  classification.
\newblock In {\em Proceedings of the Conference on Empirical Methods in Natural
  Language Processing. Conference on Empirical Methods in Natural Language
  Processing}, volume 2016, page 795. NIH Public Access, 2016.

\bibitem{zhou2016learning}
Bolei Zhou, Aditya Khosla, Agata Lapedriza, Aude Oliva, and Antonio Torralba.
\newblock Learning deep features for discriminative localization.
\newblock In {\em Proceedings of the IEEE conference on computer vision and
  pattern recognition}, pages 2921--2929, 2016.

\end{thebibliography}

\end{document}